\documentclass[10pt,twocolumn,letterpaper]{article}

\usepackage[applications]{wacv}   

\usepackage{graphicx}
\usepackage{amsmath}
\usepackage{amssymb}
\usepackage{booktabs}
\usepackage{multirow}

\usepackage{algorithm, algpseudocode}
\usepackage{comment}
\usepackage{color}
\usepackage{booktabs}
\usepackage{siunitx}
\usepackage[whole]{bxcjkjatype}
\usepackage{bm}
\input{mlpreamble}

\definecolor{wacvblue}{rgb}{0.21,0.49,0.74}
\usepackage[pagebackref,breaklinks,colorlinks,allcolors=wacvblue]{hyperref}

\newcommand\blfootnote[1]{%
  \begingroup
  \renewcommand\thefootnote{}\footnote{#1}%
  \addtocounter{footnote}{-1}%
  \endgroup  
}

\def\wacvPaperID{2096} 
\def\confName{WACV}
\def\confYear{2026}
\newcommand{\modelname}{CattleAct\xspace}

\begin{document}

\title{Interaction-via-Actions: Cattle Interaction Detection \\with Joint Learning of Action-Interaction Latent Space}

\vspace{-3mm}

\author{Ren Nakagawa$^{1\dagger}$
\quad 
Yang Yang$^{2\dagger}$
\quad 
Risa Shinoda$^{2}$
\quad 
Hiroaki Santo$^{2}$
\quad
Kenji Oyama$^{1}$ \\
Fumio Okura$^{2}$
\quad
Takenao Ohkawa$^{1}$
\\
$^1$Kobe University\qquad $^2$The University of Osaka \\
{\tt\small \{oyama,ohkawa\}@kobe-u.ac.jp}\quad{\tt\small \{yang.yang,shinoda.risa,santo.hiroaki,okura\}@ist.osaka-u.ac.jp}
}
\maketitle

\blfootnote{$^\dagger$Authors contributed equally.}


\begin{abstract} 
This paper introduces a method and application for automatically detecting behavioral interactions between grazing cattle from a single image, which is essential for smart livestock management in the cattle industry, such as for detecting estrus. Although interaction detection for humans has been actively studied, a non-trivial challenge lies in cattle interaction detection, specifically the lack of a comprehensive behavioral dataset that includes interactions, as the interactions of grazing cattle are rare events. We, therefore, propose CattleAct, a data-efficient method for interaction detection by decomposing interactions into the combinations of actions by individual cattle. Specifically, we first learn an action latent space from a large-scale cattle action dataset. Then, we embed rare interactions via the fine-tuning of the pre-trained latent space using contrastive learning, thereby constructing a unified latent space of actions and interactions. On top of the proposed method, we develop a practical working system integrating video and GPS inputs. Experiments on a commercial-scale pasture demonstrate the accurate interaction detection achieved by our method compared to the baselines. Our implementation is available at \url{https://github.com/rakawanegan/CattleAct}.
\end{abstract}

\section{Introduction}

\begin{figure}[t]
    \centering
    \includegraphics[width=\linewidth]{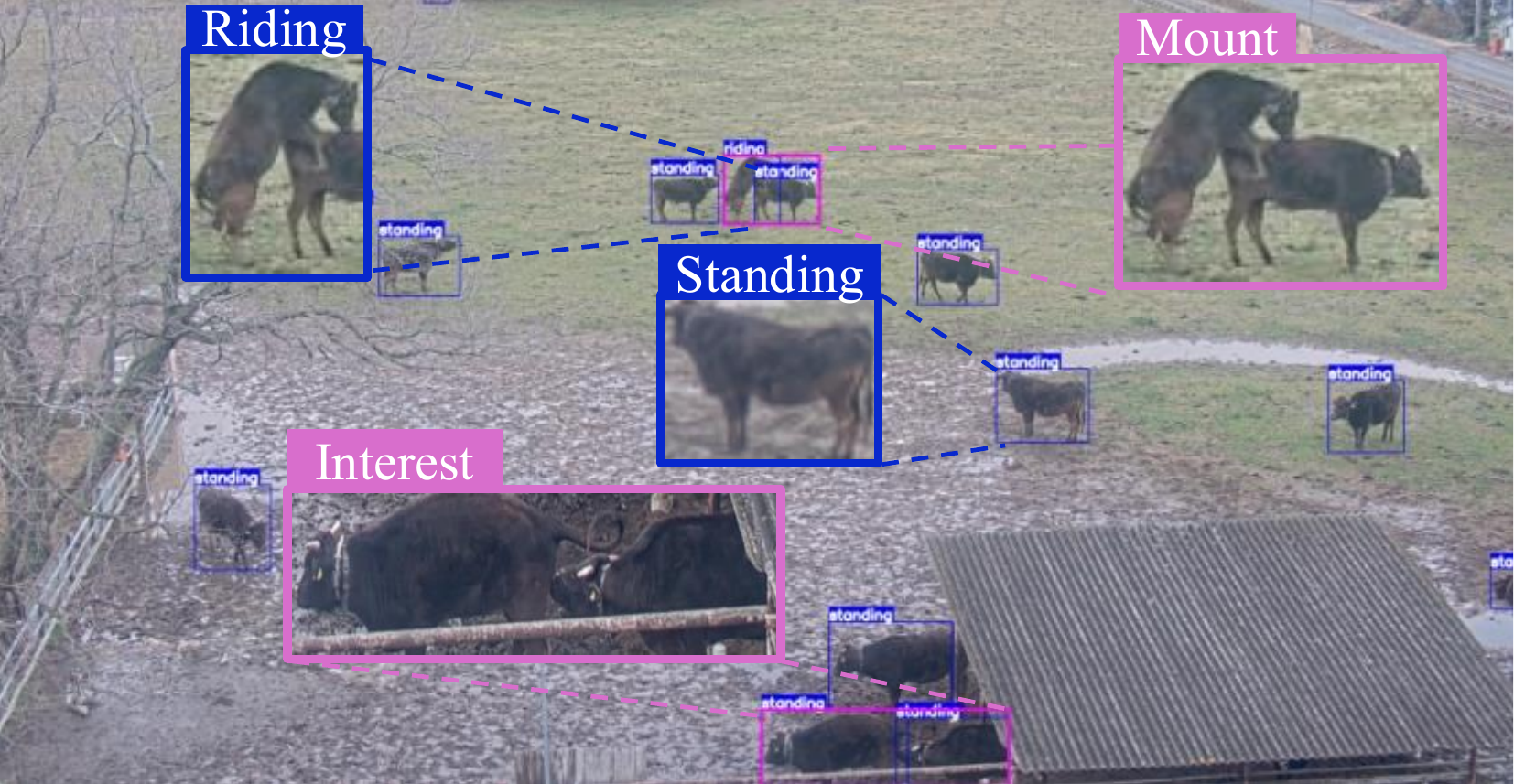}\vspace{-2mm}
    \caption{\textbf{Joint estimation of actions and interactions of cows.} Given a single-image observation, we estimate both frequent actions (\eg, standing, shown in blue) by individual cows and rare interactions (\eg, mount, shown in purple) by two cows, leveraging the joint latent space of actions and interactions.}
    \label{fig:environment}
    \vspace{-3mm}
\end{figure}

Monitoring cattle behavior is a critical component of modern livestock management, encompassing the detection of a wide range of events such as feeding, resting, social interactions, and reproductive activities, which consist of actions by single cows and interactions between two cows. Traditionally, the monitoring task has relied on farmers' visual observation, which is labor-intensive and difficult to scale in commercial farms. Automating behavioral monitoring through computer vision and sensor technologies has thus become an important research challenge.

Image or video-based action recognition has been extensively studied in the human domain~\cite{human-action-1, human-action-3}, where abundant high-quality video datasets~\cite{human-data-1, human-data-2} and well-annotated benchmarks~\cite{human-action-2} have driven the development of increasingly accurate and efficient models. These methods typically rely on spatio-temporal reasoning over dense action sequences to capture motion cues.

Extending to animal behavior analysis, many studies have employed pipelines similar to human action recognition for wildlife monitoring~\cite{Bar_2025_WACV, wildlife-1, animal-transformer-wild}, livestock management~\cite{pig-interaction, yang-scirnet,yang-triplenet, livestock-1}, and behavioral research~\cite{livestock-2,mathis2018deeplabcut}. Most methods process short video clips using models such as 3DCNNs~\cite{yang-3dcnn,animal-3dcnn} or transformer-based video encoders~\cite{Bar_2025_WACV, animal-transformer-wild} to capture visual features while incorporating pose cues~\cite{yang-scirnet,nath2019using}. 
However, in commercial farm practice, interaction events of multiple cows, such as mounting or conflicts, occur rarely compared to individual actions like grazing or lying. This long-tail distribution is inevitable in real herds and cannot be solved simply by scaling up data collection, since even continuous recording over extended periods yields only a few positive samples of interactions. Therefore, \emph{interaction} detection that effectively leverages abundant individual \emph{action} data for cows is crucial for automated livestock production. 

To address these, we propose \modelname, a data-efficient framework that decomposes interactions into the constituent actions of individual cattle, resulting in the joint estimation of actions and interactions of cows as shown in \fref{fig:environment}. 
We first construct an action latent space by pre-training on a large collection of individual action data. We then embed rare interaction cases by fine-tuning the pre-trained latent space with contrastive learning, aligning action and interaction latents within a unified space. This alignment is expected to provide mutual regularization; \ie, action features constrain interaction predictions, while interaction cues provide context to resolve ambiguities in individual actions.
While video provides temporal information, constructing large-scale video datasets with precise temporal annotations is costly. Additionally, continuously storing and processing video streams is unrealistic in real-world farm environments due to storage, computational, and communication constraints. Since the target behaviors are recognizable from single frames, we design our method for image-based inputs, which enables scalable annotation, efficient training, and practical deployment. This approach supports continuous monitoring in a cost-effective manner by leveraging lightweight inference from sampled frames.

On top of the proposed method, we develop a practical system for a production-scale pasture, applying a data augmentation method that preserves semantic information by referencing skeletal information. Experiments on a pasture of Japanese Black cattle demonstrate that aligning action and interaction embeddings significantly improves the recognition accuracy compared to baselines that model them independently or those trained on interactions alone.

\vspace{-4mm}
\paragraph{Contributions}
Our principal contributions are threefold:
\begin{itemize}
    \setlength{\parskip}{0cm}
    \setlength{\itemsep}{0cm}
    \item We introduce a novel framework that decomposes rare interactions into combinations of individual actions and aligns their embeddings in a unified feature space, enabling data-efficient training.
    \item We propose a data augmentation strategy that artificially simulates occlusion between overlapping cattle, improving the robustness in dense pasture environments.
    \item We develop and deploy a practical multimodal system that integrates video and GPS data for automatic cattle interaction detection, and demonstrate its effectiveness in a commercial pasture setting.
\end{itemize}

\begin{figure*}[t]
    \centering
    \includegraphics[width=\linewidth]{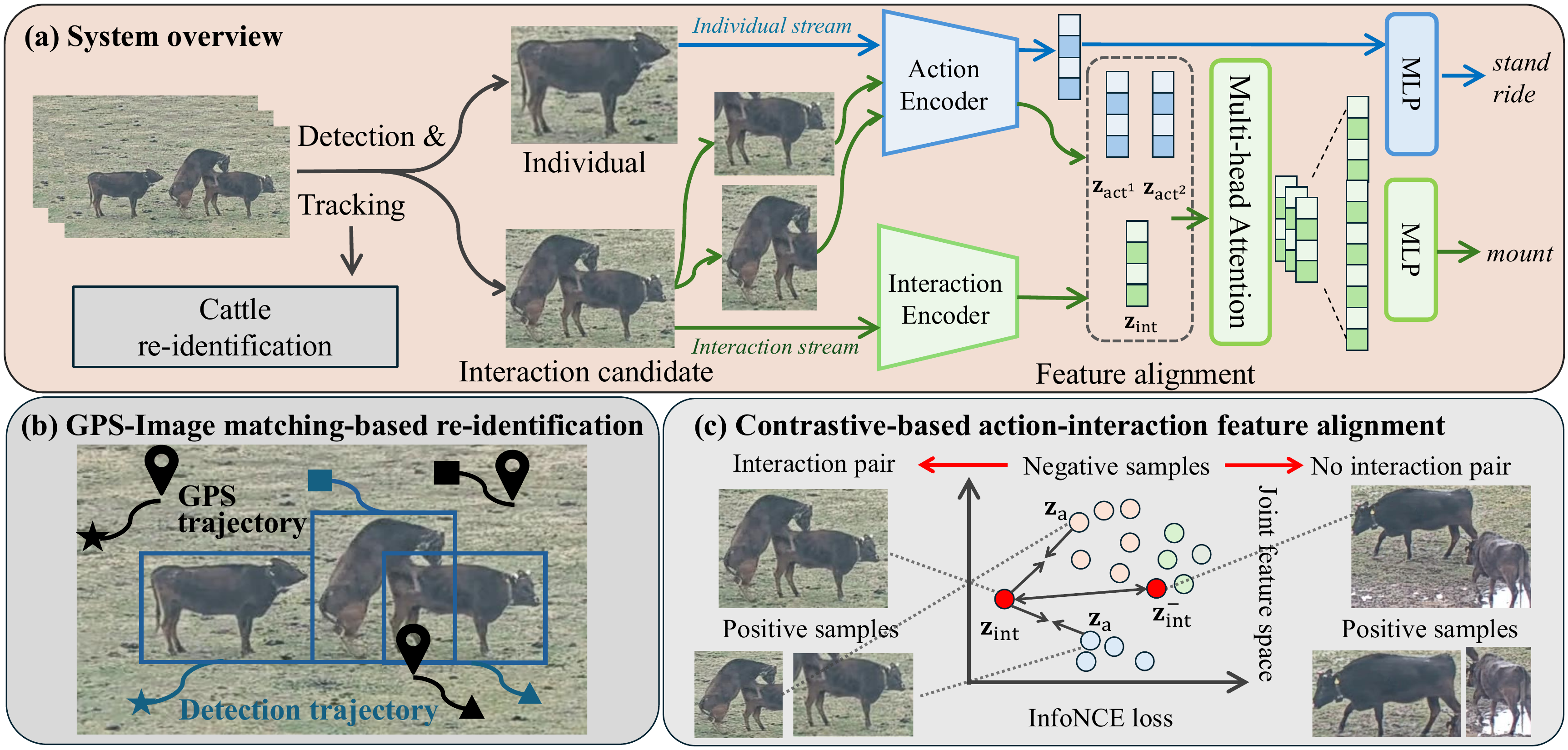}
    \caption{\textbf{Overview of \modelname}. (a) System overview: detected and tracked cows are re-identified, split into individuals and interaction candidates, and encoded by the action and interaction encoders, whose features are aligned by a multi-head attention module for final action–interaction classification.
    (b) GPS–image matching-based re-identification aligns GPS trajectories with detection trajectories to maintain consistent cattle IDs.
    (c) Contrastive-based action–interaction feature alignment pulls features of interacting pairs together while pushing apart negative and no interaction pairs in the joint latent space.} 
    \label{fig:overview}
\end{figure*}

\section{Related Work}

\paragraph{In-the-wild animal monitoring}
Automated monitoring of wild animals~\cite{Bar_2025_WACV, animal-transformer-wild} is an active research field in ecology and biodiversity conservation. Large-scale datasets such as MammalNet~\cite{chen2023mammalnet} and Snapshot Serengeti~\cite{swanson2015snapshot} have enabled the development of species classification and population estimation models at unprecedented scales. These datasets typically contain millions of images spanning diverse habitats and species, providing strong benchmarks for visual recognition under uncontrolled conditions. Unlike livestock monitoring, their primary objective is species-level identification and distribution analysis, rather than fine-grained recognition of individuals or social interactions. The imaging setups in wildlife projects are typically sparse and rely on static camera traps over broad habitats, whereas farm setups place fixed cameras to capture dense herds at closer range and larger scale.

\vspace{-5mm}
\paragraph{Livestock action detection}
In contrast, livestock monitoring focuses on individual-level analysis to improve farm management and animal welfare. Traditional approaches rely on wearable or implantable sensors~\cite{cattle-sensor} such as GPS collars, RFID tags, or accelerometers, which are reliable for coarse activity recognition (\eg, feeding, locomotion) and long-term behavioral tracking. Nevertheless, such methods are limited in their ability to capture subtle actions or complex social interactions, while also incurring costs for deployment, maintenance, and data synchronization.

Vision-based systems emerge for low-cost, scalable observation of livestock behaviors. With advances in computer vision, techniques for animal detection~\cite{yolo, maskrcnn}, tracking~\cite{yang2023track, deepsort}, and pose estimation~\cite{hr-net,lauer2022multi} have been applied to both barn and pasture environments. Early studies in controlled indoor settings~\cite{pig-interaction, cattle-indoor} concentrated on individual-level activity recognition, often employing spatio-temporal models such as 3DCNNs~\cite{yang-3dcnn}. More recent studies~\cite{yang-scirnet,yang-triplenet, yolo-interaction} have extended vision-only systems to outdoor pasture conditions, where challenges such as dense clustering, frequent occlusion, and visually uniform coats significantly degrade recognition performance~\cite{yang-scirnet}. Despite these difficulties, vision-only systems remain highly promising because they avoid the maintenance and exchange costs of heterogeneous sensors and thus hold strong potential for scalable, practical deployment.

\vspace{-5mm}
\paragraph{Livestock interaction detection}
Beyond individual action recognition, a growing body of research has focused on vision-based interaction recognition in livestock~\cite{yolo-interaction,yang-scirnet,yang-triplenet,pig-interaction}. Wang~\etal~\cite{yolo-interaction} employed a YOLO-based detector combined with attention mechanisms to identify pairwise interactions, but the approach lacked explicit modeling of relational structures. Yang~\etal~\cite{yang-triplenet} demonstrated the feasibility of interaction recognition in Japanese Black cattle through a triple-stream network, but their method still relied mainly on detection features without graph reasoning. 
To address this, SCIRNet~\cite{yang-scirnet} proposed a more advanced framework that constructed both image-based features and multiple types of inter-body graphs, which have been widely explored in human cases~\cite{li2022two}, to capture relative position cues. 
This system also considered the relation between actions and interactions, but the coupling was limited to priors pre-computed from labels.

In contrast, our approach advances interaction recognition by explicitly aligning action and interaction embeddings within a shared feature space，which encodes behavioral priors in a learnable manner and allows the two levels of behavior to directly regularize each other.

\section{Method}
The core of our method is the joint learning of frequent action and rare interaction, where interactions are decomposed into combinations of two individual actions. 
A single-cattle encoder produces action embeddings (\eg, grazing, lying, standing), while a multi-cattle encoder processes candidate pairs to produce interaction embeddings (\eg, no interaction, interest, conflict, mounting), as illustrated in \fref{fig:overview}.
By aligning action and interaction embeddings within a unified feature space, we enable data-efficient training: abundant individual action data regularizes rare interaction examples, while interaction cues provide context to resolve ambiguities in single-animal action recognition. We here suppose the images of individual cows $I_\text{act}$ and sets of cows (\ie, interaction candidates) $I_\text{int}$ are given as bounding boxes. We later describe the practical applications, including the detection and tracking of cows.

\subsection{Pre-training of Action Space} \label{sec:pretrain}
We first train an action latent space as the foundation for subsequent interaction modeling. We use a pre-trained vision transformer (ViT)~\cite{vit} backbone and fine-tune it on the cattle action dataset using a metric learning objective. 

Specifically, the action encoder is trained with a triplet loss~\cite{triplet_loss} to enforce intra-class compactness and inter-class separation in the Euclidean space, while a zero-mean regularization term keeps the embedding distribution centered and compact. This yields a discriminative latent space in which routine behaviors such as grazing, lying, and standing are well separated, providing a reliable basis for aligning interactions with their constituent actions in the next stage.
Formally, let $f_\text{act}$ denote the action encoder, and a triplet of images $(I_\text{act}, I_\text{act}^\text{p}, I_\text{act}^\text{n})$ consisting of an anchor sample $I_\text{act}$, a positive sample $I_\text{act}^\text{p}$ from the same action class, and a negative sample $I_\text{act}^\text{n}$ from a different class.
The latents are generated as
\begin{equation}
    \V{z}_\text{act}, \V{z}_\text{act}^\text{p}, \V{z}_\text{act}^\text{n} = f_\text{act}(I_\text{act}), f_\text{act}(I_\text{act}^\text{p}), f_\text{act}(I_\text{act}^\text{n}).
\end{equation}
The loss is defined using the Euclidean distance $d(\cdot)$ as
\begin{equation}
\mathcal{L}_{\text{triplet}}
= \max\!\Big(0,\ d(\V{z}_\text{act}, \V{z}_\text{act}^\text{p} ) - d(\V{z}_\text{act}, \V{z}_\text{act}^\text{n}) + \alpha \Big),
\end{equation}
where $\alpha$ is the margin hyperparameter.

\subsection{Joint Optimization of Interaction and Action} \label{sec:joint}
Given a pre-trained action encoder that provides a well-separated latent space for individual actions, this stage is to jointly optimize the representation of actions and interactions. The key motivation of joint optimization is that actions and interactions are not independent since interactions are composed of specific combinations of individual actions, which are often governed by co-occurrence and mutual exclusivity relationships (\eg, lying and grazing rarely co-occur with mounting, while mounting often co-occurs with standing and riding).

\vspace{-3mm}
\paragraph{Architecture}
Suppose we divide the interaction candidate image $I_\text{int}$ into the images containing two individual cows $I_\text{act}^1, I_\text{act}^2$.
Given an image triplet ($I_\text{int}$,$I_\text{act}^1$,$I_\text{act}^2$), the encoders produce the latents:
\begin{equation}
    \V{z}_\text{int}, \V{z}_\text{act}^1, \V{z}_\text{act}^2 = f_{\text{int}}(I_\text{int}), f_{\text{act}}(I_\text{act}^1), f_{\text{act}}(I_\text{act}^2),
\end{equation}
all in $\mathbb{R}^D$. The action and interaction latent spaces are aligned to increase mutual information. 

We apply multi-head self-attention ($\text{MHA}$) over the latents, and flatten the attended outputs to obtain the combined feature $\V{z}\in \mathbb{R}^{3\times D}$:
\begin{align}
\V{z}^\text{out} &= \mathrm{MHA}(\V{z}^\text{in}, \V{z}^\text{in}, \V{z}^\text{in})  \\
\V{z} &= \mathrm{vec}(\V{z}^\text{out}) \in \mathbb{R}^{3\times D},
\end{align}
where $\mathbf{z}^{in}=
\begin{bmatrix}
\V{z}_{\text{int}}\\
\V{z}_{\text{act}^1}\\
\V{z}_{\text{act}^2}
\end{bmatrix}\in\mathbb{R}^{3\times D}$ 
is the input latent and $\mathrm{vec}$ denotes the flatten operator. Finally, $\V{z}$ is fed into a classification layer. 

\vspace{-3mm}
\paragraph{Action-interaction feature alignment}
The goal of alignment is to maximize the mutual information between interaction embeddings and the embeddings of their decomposed actions, thereby constructing a joint feature space in which the two levels of representation are coherently organized and semantically consistent.
Specifically, we treat each interaction sample and its decomposed action crops as positive pairs, encouraging their embeddings to be pulled closer in the joint space. Negative samples are defined as mismatched pairs between an interaction and a no interaction crop. Formally, given an interaction embedding $\V{z}_{\text{int}}$, its associated action embeddings $\V{z}_\text{act}^+$ and no interaction embedding $\V{z}_\text{int}^-$, we adopt an InfoNCE~\cite{infonce} loss to align action and interaction features as
\begin{equation}
\mathcal{L}_{\text{aln}}
= -\log \frac{e^{\text{sim}(\V{z}_{\text{int}}, \V{z}_\text{act}^+)/\tau }}
{e^{\text{sim}(\V{z}_{\text{int}}, \V{z}_\text{act}^+)/\tau } 
+ \sum\limits_{j=1}^{M} e^{\text{sim}(\V{z}_{\text{int}}^{}, \V{z}_{\text{int}}^-)/\tau}},
\end{equation}
where $\tau$ is a temperature hyperparameter and $\text{sim}(\cdot)$ refers to cosine similarity.

\vspace{-3mm}
\paragraph{Interaction classification}
To mitigate class imbalance, we adopt the label-distribution-aware margin (LDAM) loss~\cite{ldam}:
\begin{equation}
\mathcal{L}_{\text{cls}}(p, y) 
= - \log 
\frac{e^{p_{y} - \Delta_{y}}}
     {e^{p_{y} - \Delta_{y}} + \sum_{j \neq y} e^{p_{j}}},
\end{equation}
where $p=[p_{1},\dots,p_{C}] \in \mathbb{R}^{C}$ are the classifier logits and $y$ is the ground-truth label.
The class-dependent margin is defined as
\begin{equation}
\Delta_{j} = \frac{C}{n_{j}^{1/4}}, \quad j \in \{1,\dots,C\},
\end{equation}
with $n_j$ the number of training samples in class $j$. 

\vspace{-3mm}
\paragraph{Overall loss}
Our proposed interaction recognition network is trained using the following total loss function:
\begin{equation}
    \mathcal{L} = \lambda_1 \mathcal{L}_{\text{aln}} + \lambda_2\mathcal{L}_{\text{cls}},
\end{equation}
where $\lambda_1$ and $\lambda_2$ are the weighting factors of two losses, respectively.

\subsection{Application} \label{sec:app}
Based on the proposed method, we introduce a practical application working on commercial-scale pastures.
Specifically, we newly developed a skeleton-aware data augmentation method and a practical cow monitoring system using video and GPS inputs.

\begin{figure}[tp]
    \centering
    \includegraphics[width=1.0\linewidth]{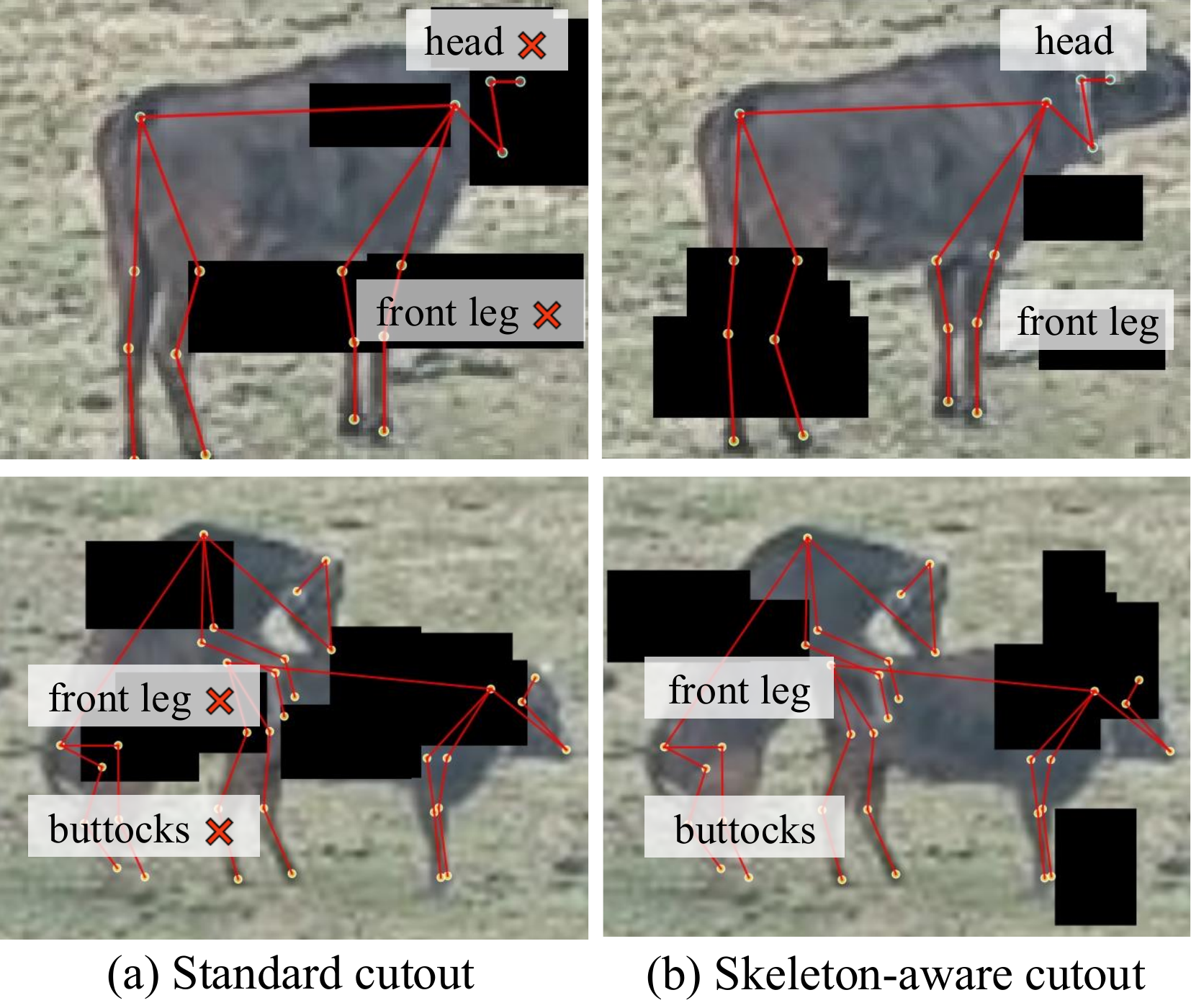}\vspace{-2mm}
    \caption{\textbf{Skeleton-aware data augmentation.} Skeleton-aware cutout enhances robustness to occlusion while selectively preserving joints vital for recognition, \ie, head and front legs for individual action recognition, and the head and torso for interaction recognition.}
    \label{fig:data_augmentation}
    \vspace{-3mm}
\end{figure}

\vspace{-3mm}
\paragraph{Skeleton-aware data augmentation}
In pastures where multiple individuals are densely active, occlusion due to overlapping individuals frequently occurs. To build a model that functions robustly under such real-world conditions, we develop a new data augmentation method by artificially simulating occlusion in the training data, as shown in \fref{fig:data_augmentation}.

While cutout \cite{cutout} is a common data augmentation method for improving robustness against occlusion, its straightforward application to images carries the risk of losing semantic information. If the model's identification accuracy improves numerically while semantic information is lost, it is highly probable that the model is not improving its real-world robustness but rather learning dataset-specific noise.

To prevent this, in this study, in addition to general data augmentation, we use a cutout-based data augmentation that ensures label invariance based on skeleton information. Specifically, information from the head and front legs is crucial for identifying individual actions, while information from the head and buttocks is important for interactions. If these relevant areas are obscured by a cutout, label invariance could be compromised. Therefore, we use skeleton information to protect these areas from being masked by Skeleton-aware cutout data augmentation.

\vspace{-3mm}
\paragraph{System design}
Based on the proposed method, we develop a practical system working on commercial-scale pastures using videos and GPS. Our system takes as input synchronized video streams from fixed pasture cameras and GPS trajectories from wearable collars, and outputs both individual action labels and pairwise interaction labels. Additionally, GPS data is utilized exclusively for cattle re-identification, while behavior labels are predicted solely based on visual information, as illustrated in \fref{fig:overview}.

From the input video, we first apply object detection using YOLO~\cite{yolo} and tracking using DeepSORT~\cite{deepsort} to obtain frame-level bounding boxes of individual cattle. These cropped sequences are further processed by a pose estimator to extract 2D skeleton sequences. To construct candidate interaction pairs, we merge overlapping bounding boxes that satisfy a spatial threshold. In parallel, GPS trajectories provide coarse position information for each animal. We project GPS coordinates into the image plane using a perspective transform and apply a combinatorial matching algorithm to associate tracklets with GPS identities. This ensures robust re-identification of individuals, especially in crowded scenes, and facilitates downstream herd management.\looseness=-1


\section{Experiments}
We evaluate the effectiveness of our proposed method on the task of behavior recognition for cattle from both quantitative and qualitative aspects. The evaluation is conducted from two perspectives: individual action recognition and interaction recognition between individuals. For interaction recognition, we thoroughly evaluate our method with several baselines. For the evaluation of action recognition, the main research question is whether our joint learning contributes to the action recognition accuracy; therefore, we evaluate the action recognition in the ablation study.

\subsection{Experiment Setup}

\paragraph{Evaluation metrics} For the action and interaction recognition tasks, we adopt accuracy and F1-score as evaluation metrics. In particular, to evaluate the recognition performance of rare events such as estrus signs, we also use the macro F1-score, which can account for class imbalance.

\vspace{-3mm}
\paragraph{Baselines}
We compare our method with the following baseline methods.
\begin{itemize}
\vspace{2mm}
\item \textbf{Animal-CLIP}~\cite{animalclip}: Animal CLIP is a zero-shot classifier that takes a cropped image as input. We utilized a model pre-trained on the Mammal Net dataset, which is designed for animal behavior recognition and includes interaction labels. The prompts were manually created following the specified format. 
For instance, the prompt for ``mounting" is defined as ``\textit{This shows the behavior of mounting. This is a specific action, often related to dominance or reproduction, where one cattle physically climbs onto the back of another animal. The front legs of the mounting cattle are typically on the other's rear quarters}''.
\vspace{2mm}
\item \textbf{E-YOLO}~\cite{yolo-interaction}: E-YOLO is a model that concurrently performs the tasks of object detection and classification from a single input frame. 
The model was trained for 300 epochs with an input image size of $640 \times 640$ pixels, consistent with the reference literature. 
The loss function gains were set to $15$ for the NWD loss~\cite{nwd_loss}, $3$ for the Distribution Focal Loss~\cite{focal_loss}, and $1$ for the BCE loss. 
Following the methodology of the E-YOLO paper, we did not label individual actions, focusing exclusively on labeling interactions.
In this evaluation, to assess the performance specifically as an identification task, misclassifications as ``background'' were not penalized.
\vspace{2mm}
\item \textbf{SCIRNet}~\cite{yang-scirnet}: SCIRNet is a straightforward baseline designed for cattle interaction recognition that utilizes both images and their corresponding skeleton pose as input. This model is designed to jointly process visual and skeletal information to make a final prediction. They used ViT and STGCN~\cite{st_gcn} for image feature extraction and graph feature extraction, respectively. We use HRNet~\cite{hr-net} pretrained on the Animal Kingdom dataset~\cite{animal-kingdom} to extract the cattle skeleton.
sequence of relative coordinates with three channels: $x$, $y$ and a confidence score. Features extracted from each backbone are fused via an attention mechanism to produce the final classification.
\end{itemize}

\begin{table}[t]
\centering
\caption{\textbf{Categories and sample counts for behaviors.}}
\label{tab:all_behaviors}
 \resizebox{\linewidth}{!}{
    \begin{tabular}{lrr|lrr}
    \toprule
    \textbf{Category} & \textbf{Images} & \textbf{Actions} &\textbf{Category} & \textbf{Images} & \textbf{Actions}\\
    \midrule
    \multicolumn{3}{l|}{\textit{Individual Behaviors}} & \multicolumn{3}{l}{\textit{Interactions}} \\
    \quad grazing & 2209 & 2209 &   \quad no interaction & 3637 & 1110 \\
    \quad standing & 816 & 816 &    \quad interest & 1379 & 254 \\
    \quad lying & 319 & 319 &       \quad conflict & 178 & 21\\
    \quad riding & 165 & 13 &      \quad mount & 117 & 22 \\
    \midrule
    \quad \textbf{Subtotal} & \textbf{3509} & \textbf{3357} & \quad \textbf{Subtotal} & \textbf{5311} & \textbf{1407}\\
    \bottomrule
    \end{tabular}
}
 \vspace{-3mm}
\end{table}

\begin{figure*}[t]
    \centering
    \includegraphics[width=\linewidth]{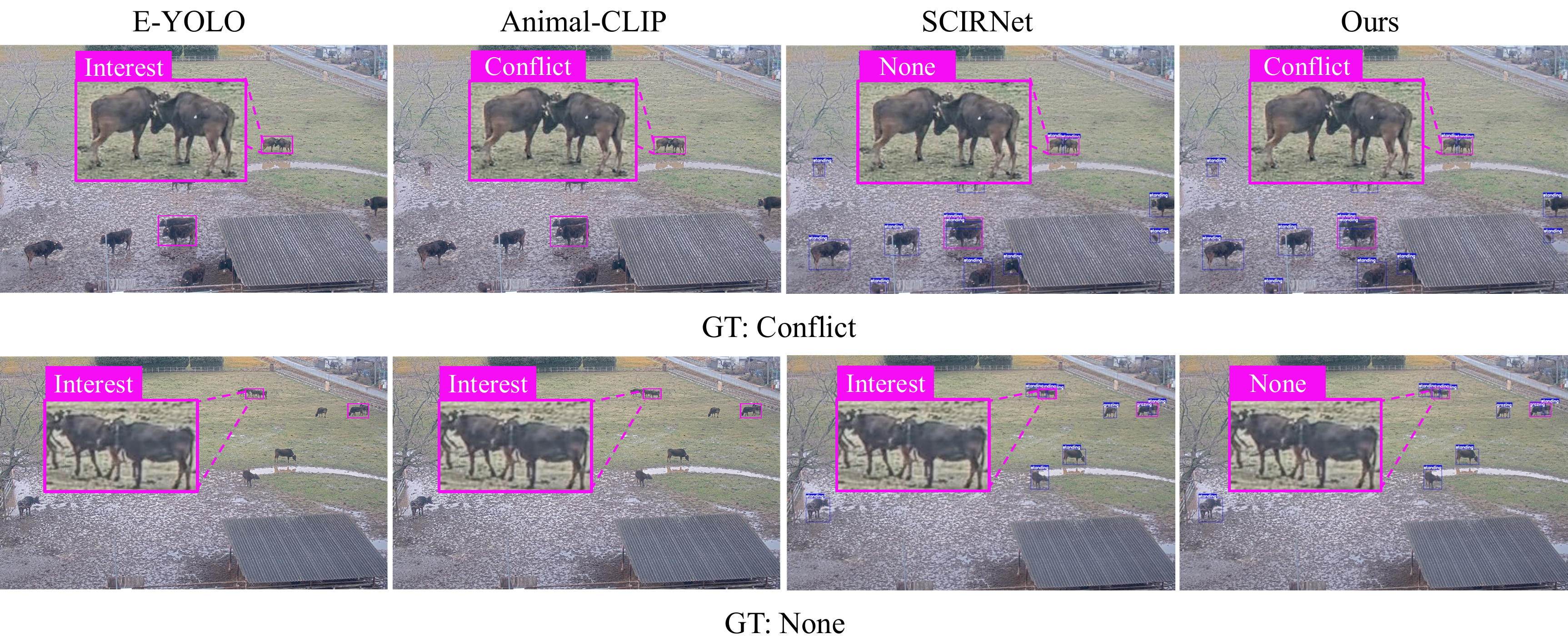}\vspace{-2mm}
    \caption{\textbf{Visual comparisons.} Our method accurately detects both actions and interactions from real-world pasture images, compared to baseline methods.}
    \label{fig:visual}
\end{figure*}

\vspace{-3mm}
\paragraph{Dataset}
In this study, we constructed a proprietary dataset targeting approximately 30 cattle grazed. This dataset is characterized by its collection within a commercial-scale grazing environment, simulating actual operational conditions. Data were acquired on 12 non-consecutive days between 2019 and 2025 from two fixed 4K cameras (30 FPS) installed on a silo, providing multi-angle overhead views of the pasture. To create the image dataset for annotation, frames were extracted from the video footage at one-second intervals, and individual cattle were automatically detected and cropped from these frames.

From the collected images, we annotated individual behaviors and interactions. The breakdown of categories and the number of samples is shown in \Tref{tab:all_behaviors}. It should be noted that the sampling methods for individual behaviors and interactions differ. For individual behaviors, one frame is sampled from a sequence of actions. In contrast, for interactions, frames are extracted at one-second intervals from a sequence of actions. This difference in sampling frequency results in a larger number of images for the interaction categories. For the purpose of model performance evaluation, the interaction dataset was partitioned into 3,704 samples for training, 434 for validation, and 1,173 for testing. These interactions serve as crucial indicators for detecting signs of estrus. Notably, as shown in the table, decisive events such as mounting behaviors (riding and mounting) constitute a very small fraction of the entire dataset, which results in a significant class imbalance.
\vspace{3mm}
\paragraph{Implementation details}
We utilized ViT~\cite{vit} and CNN as the backbones for action and interaction feature extraction, respectively. This choice is motivated by two considerations: (1) the use of large convolutional kernels, in practice $32 \times32$, can help capture the global feature of the interaction area; (2) interaction samples are relatively scarce compared to individual action data, so a CNN with fewer parameters is less prone to overfitting than a ViT. For both action crop and interaction crop, they are cropped and resized to $224$ $\times$ $224$ for training and testing. Random horizontal flip and RandAugment~\cite{randaugment} were also deployed as general data augmentation. The output dimension $D$ of both encoders was $256$. For action pre-training, the training parameters were: $50$ training epochs, initial learning rate $1 \times 10^{-5}$, margin $\alpha=0.5$. For interaction training, the training parameters were: $50$ training epochs, initial learning rate $1 \times 10^{-5}$, temperature hyperparameter $\tau=0.03$, loss weight $\lambda_1 = 1$ and $\lambda_2$ linearly decayed from $0.1$ to $0$.

\vspace{3mm}
\subsection{Results}

\begin{table}[tp]
\centering
\caption{\textbf{Performance comparison of interaction recognition.} }
\vspace{-2mm}
\label{tab:performance_comparison}
 \resizebox{\linewidth}{!}{
    \begin{tabular}{l|cc|c}
    \hline
    Method & Accuracy & F1-score & Modality \\
    \hline
    Animal-CLIP~\cite{animalclip} & 66.0 & 35.8 & Image \& Text\\
    E-YOLO~\cite{yolo-interaction} & 70.8 & 48.1 & Image-only  \\ 
    SCIRNet~\cite{yang-scirnet} & 73.7 & 44.9 & Image \& Graph \\
    Ours & \textbf{83.6} & \textbf{67.2} & Image-only\\
    \hline
    \end{tabular}
}
\end{table}

\paragraph{Main results}
\Tref{tab:performance_comparison} shows the recognition results for interactions between cows. 
Since ``mounting'', a direct sign of estrus, is extremely rare, the baseline methods performed poorly. In contrast, our method, which combines pre-training and data augmentation, significantly improves the F1-score. Also, for ``interest'', a potential preceding indicator of estrus, and ``conflict'', an indicator of herd health, our method demonstrated stable and high recognition performance, significantly surpassing the baselines. \Fref{fig:visual} shows visual examples, indicating the better recognition for both accuracy and interaction in real-world pasture images.

\begin{table}[t]
\centering
\caption{\textbf{Per-class accuracy and F1-score comparison across four methods.}}
\label{tab:perclass}
 \resizebox{\linewidth}{!}{
\begin{tabular}{lcccccccc}
\toprule
\multirow{2}{*}{Class} & 
\multicolumn{2}{c}{SCIRNet} & 
\multicolumn{2}{c}{E-YOLO} & 
\multicolumn{2}{c}{Animal-CLIP} & 
\multicolumn{2}{c}{Ours} \\
\cmidrule(lr){2-3} \cmidrule(lr){4-5} \cmidrule(lr){6-7} \cmidrule(lr){8-9}
& Acc. & F1 & Acc. & F1 & Acc. & F1 & Acc. & F1 \\
\midrule
None     &  45.8 & 51.6 & 85.5 & 91.8 & 72.9 & 81.3 & 86.6 & 91.7 \\
Interest &  90.2 & 17.3 & 90.0 & 17.0 & 86.8 & 17.1 & 88.5 & 41.1 \\
Conflict &  40.2 & 15.9 & 94.4 & 0.0 & 78.8 & 33.0 & 92.7 & 45.6 \\
Mount    &  99.7 & 93.9 & 99.6 & 91.2 & 93.6 & 11.8 & 99.5 & 90.3 \\
\midrule
\textbf{Avg.} & 69.0 & 44.51 & 92.4 & 50.0 & 83.0 & 35.8 & 91.8 & 67.2 \\
\bottomrule
\end{tabular}
}
\end{table}

\vspace{-3mm}
\paragraph{Per-class interaction recognition analysis}
For real-world deployment, metrics directly linked to system reliability are crucial. The first is the ability to correctly identify ``no interaction'' scenarios, thereby suppressing false positives. The second is the high recognition accuracy for ``mounting,'' a key estrous behavior in reproductive management. \Tref{tab:perclass} summarizes the per-class performance of interaction recognition. While Animal-CLIP excels in distinguishing the presence or absence of interaction and E-YOLO shows strength in mounting recognition, our method maintains a high level of performance on both metrics. This demonstrates a clear advantage in the balanced performance required for practical use.

\begin{figure}[tp]
    \centering
    \includegraphics[width=1.0\linewidth]{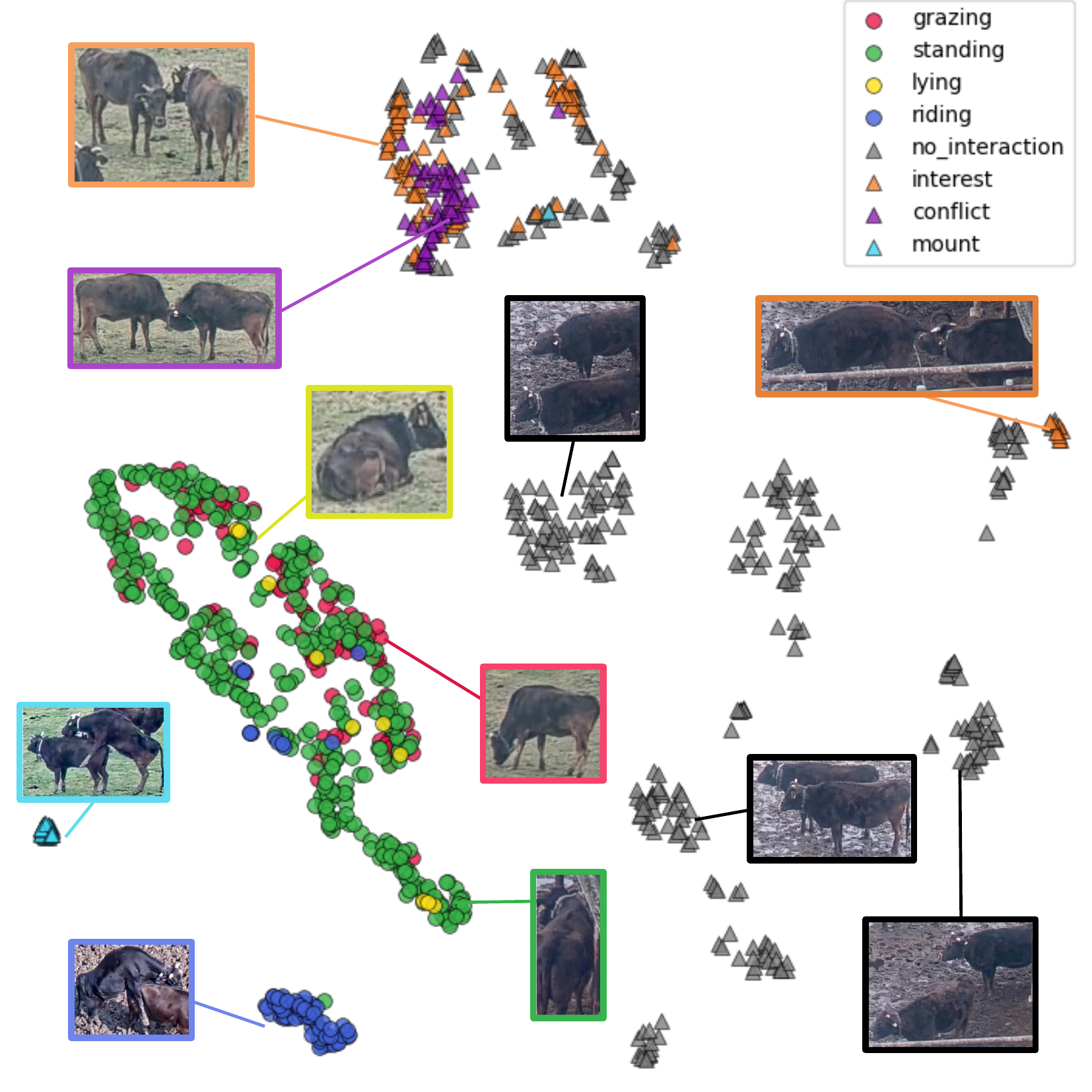}
    \caption{\textbf{t-SNE visualization of the feature space.} Each point represents a sample, classified by markers (circles: individual action, triangles: interactions), and colors are class labels.}
    \label{fig:latent space}
\end{figure}

\vspace{-3mm}
\paragraph{Feature space visualization}
To visualize the effect of our proposed learning process on the feature space, we used t-SNE \cite{t-sne} to compress the distribution of feature vectors into two dimensions, as shown in \fref{fig:latent space}.
The visualization demonstrates that behaviors such as ``mounting'' and ``riding'' are distinctly separated into different clusters, while the ``no interaction'' samples decompose into multiple clusters, reflecting differences in orientation and relative positioning.
Moreover, the individual action ``riding'' is located close to the interaction class ``mounting'', suggesting that the feature space can represent their semantic relatedness. This supports our design of aligning interaction and action representations, as the feature space organizes related behaviors across semantic levels.

\begin{table}[t]
\caption{\textbf{Ablation studies on each component.} We analyze the impact of each component by removing it from our full model. All values are presented in the format of mean ± standard deviation.}
\vspace{-2mm}
\centering
\resizebox{1.0\linewidth}{!}{
    \begin{tabular}{l|cc}
    \toprule
    Method & Accuracy  $\uparrow$ & F1-score $\uparrow$ \\
    \midrule
    \textbf{Full model (ours)} & \textbf{84.0 ± 5.4} & \textbf{67.1 ± 4.6} \\
    \midrule
    \textit{Ablation analysis} & & \\
    \quad w/o action pretraining & 83.0 ± 4.5 & 63.5 ± 7.6 \\
    \quad w/o action-interaction alignment loss & 84.3 ± 4.1 & 65.5 ± 7.2 \\
    \quad w/o skeleton-aware cutout$^\dagger$ & 77.3 ± 6.7 & 61.1 ± 8.6 \\
    \bottomrule
    \multicolumn{3}{l}{$^\dagger$Implementation using standard cutout augmentation.}
    \end{tabular}
}
\label{tab:ablation_components}
\vspace{-3mm}
\end{table}

\subsection{Ablation Study}

\paragraph{Ablation of key modules}
To validate the effectiveness of our proposed method, we conducted an ablation study as shown in \Tref{tab:ablation_components}. We specifically compared the performance under two conditions: (1) without pre-training with an individual action space and (2) without skeleton-aware cutout data augmentation. The results are from five independent trials, each with a different random seed. 
The largest performance drop of 8.4\% occurred when omitting the skeleton-aware cutout data augmentation, which also led to increased variance and unstable training. This result suggests that effective data augmentation is crucial for learning rare classes. The performance also degraded significantly when pre-training with an individual action space was omitted, suggesting that each component of our proposed method is essential for enhancing learning stability.

\begin{table}[t]
\caption{\textbf{Ablation studies on action recognition.}}
\vspace{-2mm}
\centering
\resizebox{0.8\linewidth}{!}{
    \begin{tabular}{l|cc}
    \toprule
    Method & Accuracy  $\uparrow$ & F1-score $\uparrow$ \\
    \midrule
    W/o skeleton-aware cutout & 95.6 & 92.6 \\
    \textbf{Full model (ours)} & \textbf{96.5} & \textbf{94.5}  \\
    \midrule
    \end{tabular}
}
\label{tab:ablation_action}
\end{table}

\vspace{-3mm}
\paragraph{Effect on action recognition accuracy} \Tref{tab:ablation_action} shows the action recognition results with and without skeleton-aware cutout. We find that incorporating skeleton-aware cutout consistently improves performance, which verifies that preserving key body joints during augmentation provides more informative supervision. The accuracy of individual action recognition is considerably high, achieving a practical level.

\begin{figure}
    \centering
    \includegraphics[width=1.0\linewidth]{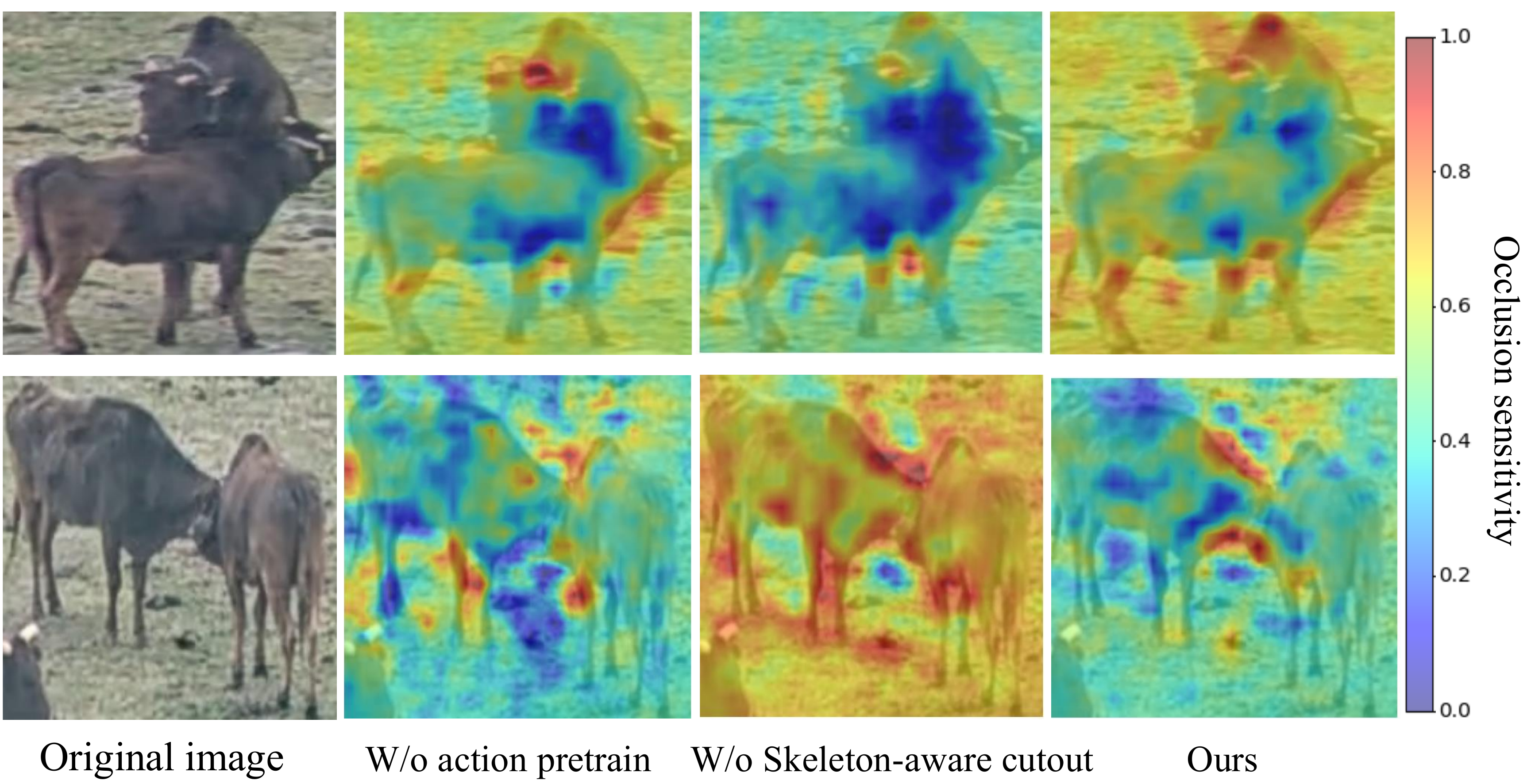}\vspace{-2mm}
    \caption{Comparison of occlusion sensitivity with and without skeleton-aware cutout data augmentation. Occlusion sensitivity involves moving a mask simulating occlusion across the image to visualize the stability of the inference results.}
    \label{fig:occlusion_sensibility_cutout_comparison}
\end{figure}

\vspace{-3mm}
\paragraph{Effect on occlusion sensitivity}
We evaluated the model's robustness against occlusion, a common occurrence in actual grazing environments. Since the overlap between cattle is a primary cause of performance degradation, we verified the model's stability through occlusion sensitivity analysis.
We first compute the target class score on the unoccluded image as the baseline. Then, we occlude local patches of the image and measure score drops. The occlusion sensitivity value at each patch is defined as the score drop, and the values are aggregated into a heatmap to visualize regions most critical for the prediction.
As shown in \fref{fig:occlusion_sensibility_cutout_comparison}, in an ablation model where key components of our method were removed, the inference for ``mounting'' did not change even when indispensable regions, such as the buttocks of the mounting cow and the head of the mounted cow, were masked. This suggests that the model is not focusing on the appropriate feature regions as a basis for its judgment. Furthermore, the model without the skeleton-aware cutout component exhibited unstable inference for the ``conflict'' action. These results indicate that the proposed method, particularly the skeleton-aware cutout, ensures the stability of inference and the validity of the judgment basis under occlusion, thereby significantly contributing to its robustness in real-world applications.

\subsection{Failure Cases}
\begin{figure}[t]
    \centering
    \includegraphics[width=\linewidth]{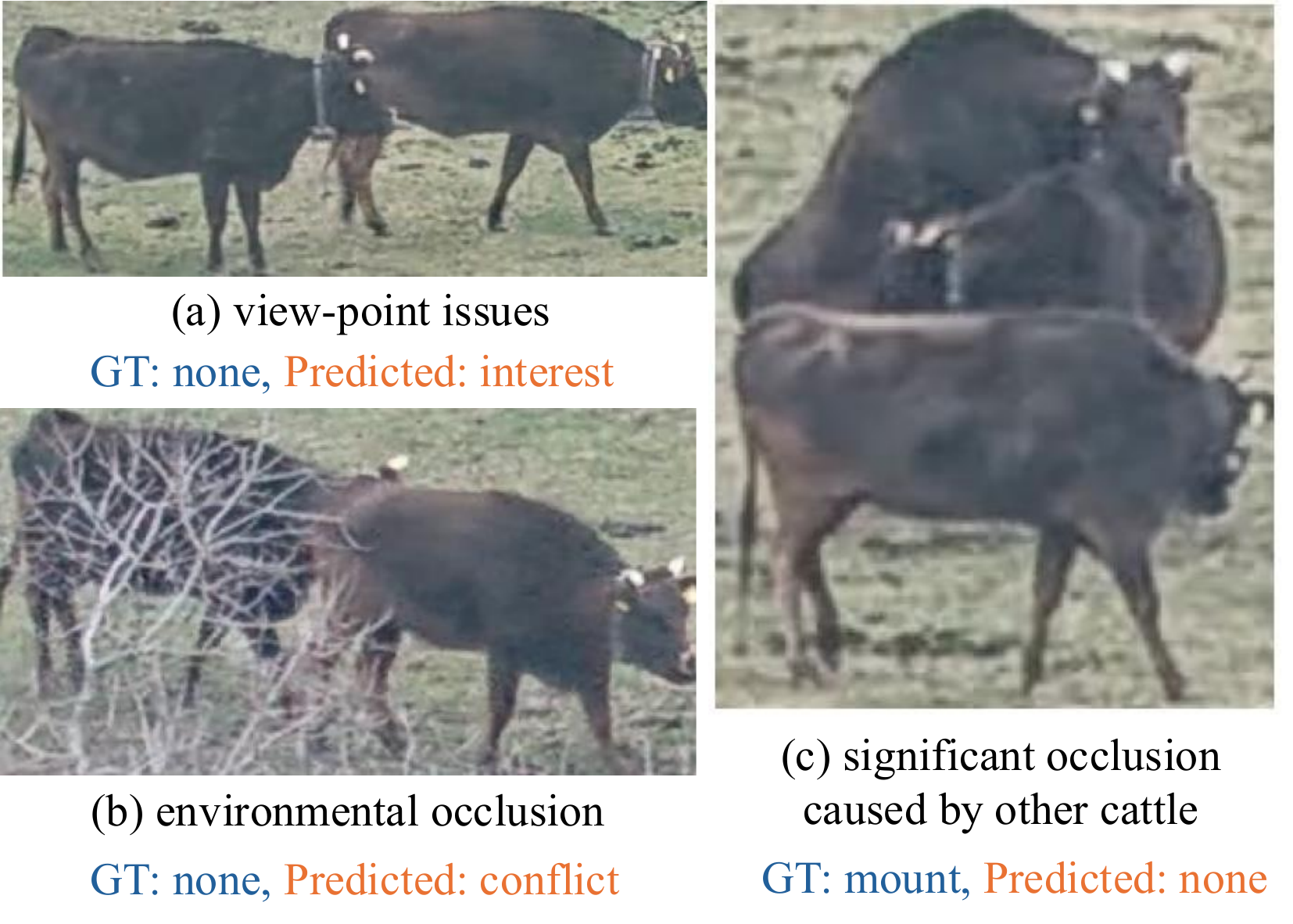}\vspace{-2mm}
    \caption{Failure cases.}
\vspace{-3mm}
    \label{fig:failure}
\end{figure}

Failure cases of the proposed method are largely categorized into three types as illustrated in \fref{fig:failure}: (a) viewpoint issues, (b) environmental occlusion, and (c) significant occlusion by other cattle. Viewpoint issues also cause misidentification. When multiple cattle are positioned so that they appear to be in close contact from a specific viewpoint, misidentification occurs. This leads to them being incorrectly judged as being in contact despite being in a non-contact state, resulting in the misclassifications of ``no interaction'' state as ``interest'' or ``conflict''. Occlusions by other objects (\eg, trees seen in \fref{fig:failure}(b)) are a major cause of the misclassification. Similarly, significant occlusion by other cattle can cause misidentification (see \fref{fig:failure}(c)).

\section{Conclusions}
We presented \modelname, a framework that jointly learns action and interaction representations for cattle interaction recognition. By aligning the two semantic levels in a shared feature space, our method improves robustness to occlusion and ambiguity compared to treating them independently. Importantly, the approach is well-suited to real-world farm scenarios, where data quality is relatively low, and interaction cases are scarce. Leveraging action–interaction structure provides strong regularization in such small-data settings. 
This contributes to practical, low-cost, vision-based livestock monitoring, supporting improved animal welfare and farm management.

\vspace{-3mm}
\paragraph{Limitations}
Despite its effectiveness, \modelname is still challenged by adverse farm conditions: extreme weather can degrade image quality, dense clustering leads to heavy occlusion, and ambiguous viewpoints hinder accurate recognition. Future work may integrate complementary sensing to improve robustness.

\section*{Acknowledgments}
This work was supported in part by the JSPS KAKENHI JP23H05491, JP25K03140, JP21H04914, and JST FOREST JPMJFR206F.

{
    \small
    \bibliographystyle{ieeenat_fullname}
    \bibliography{main}
}

\clearpage
\begin{appendix}
\maketitlesupplementary

\appendix

\begin{appendix}

\title{Interaction-via-Actions: Cattle Interaction Detection \\with Joint Learning of Action-Interaction Latent Space}




\section{Dataset Details} \label{supsec:dataset}

This section describes the data collection environment and the definition of action labels.

\begin{figure}[t]
    \centering
    \includegraphics[width=\linewidth]{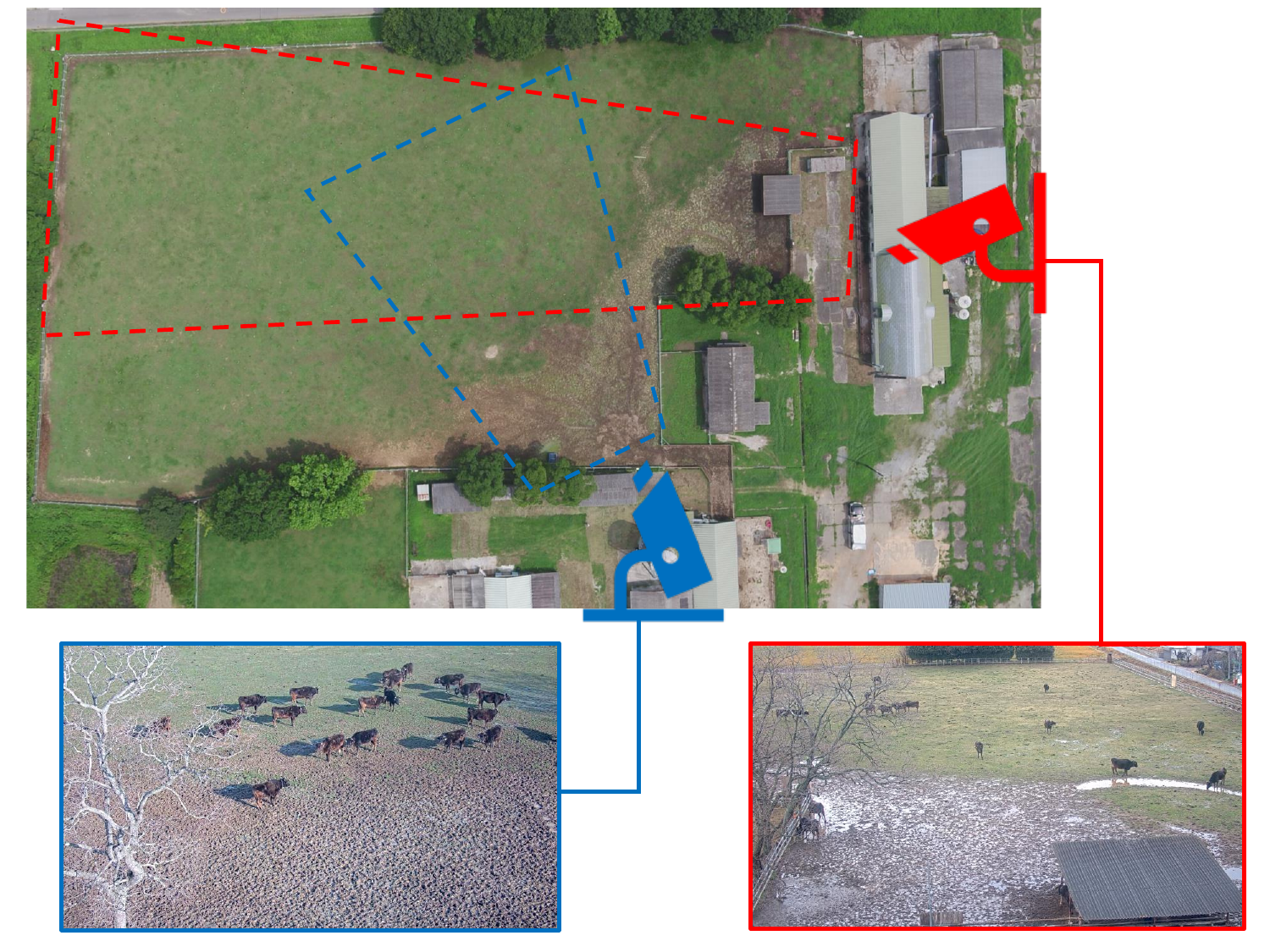}
    \caption{Data collection environment and camera configurations.
    Top: Aerial view of the pasture showing the overlapping fields of view of the two surveillance cameras.
    The red and blue dashed lines indicate the coverage areas of each camera.
    Bottom: Sample frames from each viewpoint, illustrating diversity in object scale and occlusion.}
    \label{fig:camera_view}
\end{figure}

\subsection{Environmental Settings and Data Split}
Our dataset was collected in a real-world pasture. As shown in Fig.~\ref{fig:camera_view}, a multi-camera setup was employed to enable the learning of behavioral features independent of specific camera angles.

\paragraph{Camera Angles and GPS Synchronization}
A key challenge is aligning GPS location data with visual data.
The training set utilizes video sequences from multiple angles to facilitate view-invariant feature learning.
The test set, however, is restricted to a single view (the camera indicated in red in the figure).
This view enables the most precise matching between GPS coordinates and bounding boxes, ensuring the performance required for practical application.

\begin{figure}[t]
    \centering
    \includegraphics[width=\linewidth]{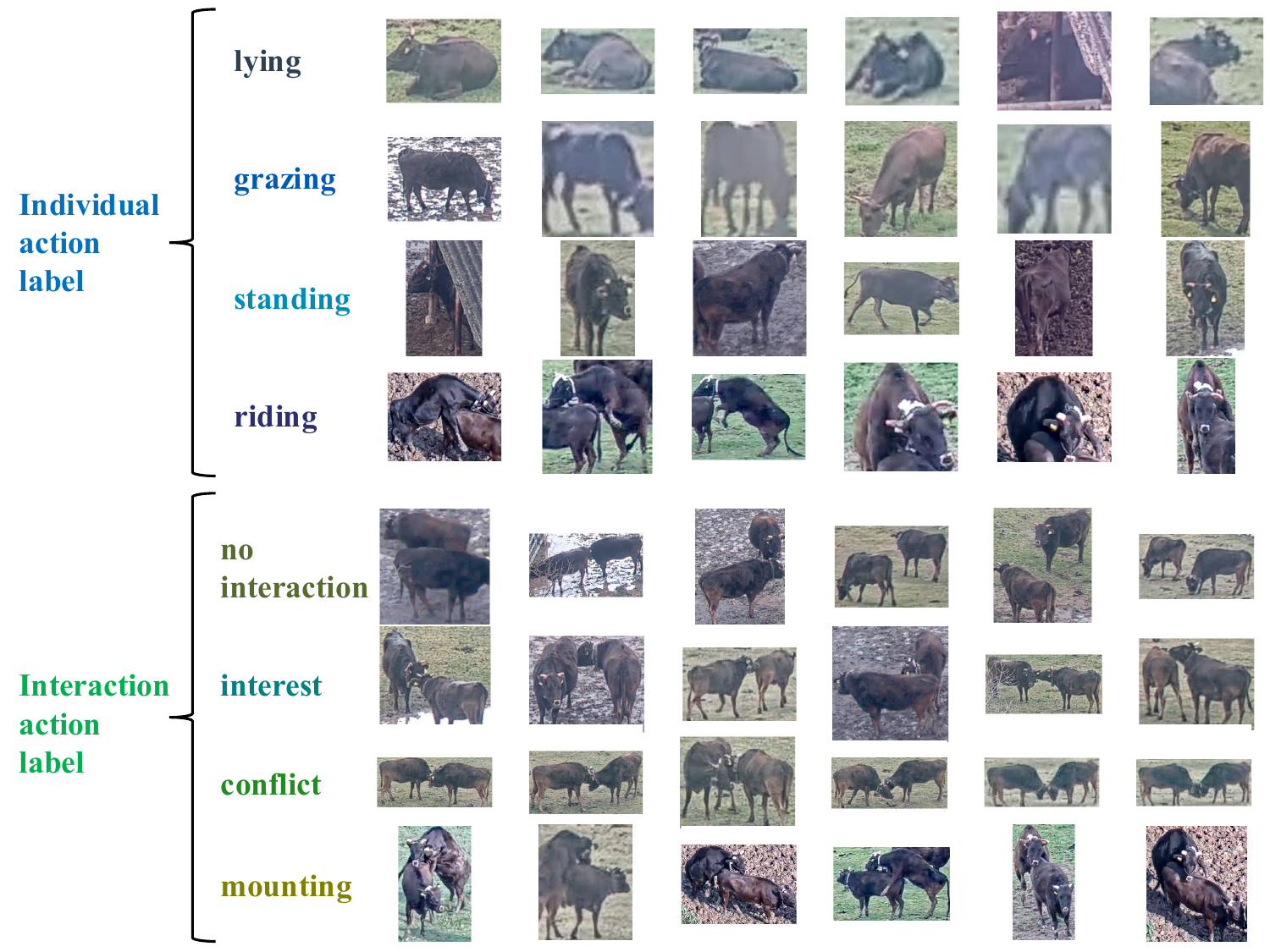}
    \caption{sample of cattle behavior labels.
    Top: Individual action labels. States of single subjects (e.g., \textit{lying, grazing, standing, riding}).
    Bottom: Interaction action labels. Relationships between multiple subjects (e.g., \textit{no interaction, interest, conflict, mounting}). High-level interactions often co-occur with specific individual actions.}
    \label{fig:dataset_sample}
\end{figure}

\subsection{Action Label Definitions}
As shown in Fig.~\ref{fig:dataset_sample}, action labels are classified into two groups.

\begin{itemize}
    \item Individual Action Labels: Physical state of the cattle. \textit{Lying} (resting), \textit{Grazing} (eating), \textit{Standing} (idle or moving), and \textit{Riding} (individual motion associated with mounting).
    \item Interaction Action Labels: Social relationships between cattle. \textit{No Interaction}, \textit{Interest}, \textit{Conflict}, and \textit{Mounting}.
\end{itemize}
These annotations aim to enable the understanding of herd dynamics beyond simple object detection.

\end{appendix}
\end{appendix}
\end{document}